\documentclass[conference]{IEEEtran}
\IEEEoverridecommandlockouts
\usepackage[numbers,sort&compress]{natbib}
\usepackage{amsmath,amssymb,amsfonts}
\usepackage{algorithmic}
\usepackage{graphicx}
\usepackage{textcomp}
\usepackage{xcolor}
\usepackage{multirow}
\usepackage{algorithm}

\usepackage{hyperref}
\hypersetup{colorlinks=true,linkcolor=blue,urlcolor=blue}

\def\BibTeX{{\rm B\kern-.05em{\sc i\kern-.025em b}\kern-.08em
    T\kern-.1667em\lower.7ex\hbox{E}\kern-.125emX}}
\begin{document}

\title{Improving Generalization of Medical Image Registration Foundation Model\\
{\footnotesize 
}
}

\author{\IEEEauthorblockN{Jing Hu$^1$, Kaiwei Yu$^1$, Hongjiang Xian$^1$, Shu Hu$^2$, Xin Wang$^{3 *}$\thanks{*Corresponding author}}
\IEEEauthorblockA{\textit{ $^1$Chengdu University of Information Technology, China} \\
\textit{ $^2$Purdue University, USA}\\
\textit{ $^3$University at Albany, State University of New York, USA}\\
jing\_hu09@163.com, kaiwei\_yu13@163.com, 2515318083@qq.com, hu968@purdue.edu, xwang56@albany.edu}
}

\maketitle

\begin{abstract}
—
Deformable registration is a fundamental task in medical image processing, aiming to achieve precise alignment by establishing nonlinear correspondences between images. Traditional methods offer good adaptability and interpretability but are limited by computational efficiency. Although deep learning approaches have significantly improved registration speed and accuracy, they often lack flexibility and generalizability across different datasets and tasks. In recent years, foundation models have emerged as a promising direction, leveraging large and diverse datasets to learn universal features and transformation patterns for image registration, thus demonstrating strong cross-task transferability. However, these models still face challenges in generalization and robustness when encountering novel anatomical structures, varying imaging conditions, or unseen modalities. To address these limitations, this paper incorporates Sharpness-Aware Minimization (SAM) into foundation models to enhance their generalization and robustness in medical image registration. By optimizing the flatness of the loss landscape, SAM improves model stability across diverse data distributions and strengthens its ability to handle complex clinical scenarios. Experimental results show that foundation models integrated with SAM achieve significant improvements in cross-dataset registration performance, offering new insights for the advancement of medical image registration technology. Our code is available at \href{https://github.com/Promise13/fm_sam}{https://github.com/Promise13/fm\_sam}.

\end{abstract}


\section{Introduction}

Deformable registration is a fundamental task in various medical image studies and has been a popular research topic for decades \cite{ashburner2007fast,dalca2016patch,glocker2008dense,thirion1998image,zitova2003image,maintz1998survey,hu2023attention,luo2022stochastic}. The goal is to establish a complex nonlinear correspondence between a pair of images. Various traditional medical image registration methods search for optimal transformation parameters in the parameter space through optimization algorithms to maximize similarity metrics, typically using rigid and elastic transformations \cite{kybic2003fast,bajcsy1989multiresolution,shen2002hammer}, and feature point- \cite{liu2011simple,thirion1996new} or intensity-based registration \cite{klein2009elastix,rohde2003adaptive} techniques. These methods have a solid theoretical foundation and high interpretability, which led to their widespread use in the early stages when computational resources were limited. Additionally, due to their reliance on anatomical features and geometric transformations, these methods achieve high accuracy in standardized feature tasks. However, because they depend on precise feature extraction and matching, they suffer from high computational complexity and long registration times \cite{modat2010fast,modat2014global}.

With the development of deep learning technology \cite{lecun2015deep,goodfellow2016deep,huang2025diffusion,wang2025graph,wang2025fg,ren2024improving,lin2025ai,fu2024vb,chen2024self,wang2024u,sun2024repmedgraf,zhu2024cgd,zhu2024qrnn,qiu2024enhancing,peng2024uncertainty,yang2024llm,qiu2024real,chen2024masked,lin2024detecting,fan2024efficient,hu2024fairness,fan2024synthesizing,hu2024umednerf,zhang2024x,xiang2023rmbench,yang2023improving,wang2023deep,hu2024outlier,fan2023attacking,ju2024improving,hu2023attention,lu2023attention,xie2023attacking,li2023ntire,chen2023harnessing,hu2023rank,hu2022distributionally}, supervised and unsupervised learning-based methods \cite{cao2017deformable,yang2017fast,balakrishnan2019voxelmorph,sokooti2017nonrigid} that use deep neural networks to automatically extract features and learn transformations have significantly improved registration efficiency \cite{de2019deep,rohe2017svf}. In many medical image registration tasks, deep learning-based methods have already surpassed traditional methods in terms of accuracy and speed. However, deep learning-based methods require task-specific registration networks, which makes them much less flexible compared to traditional image optimization algorithms \cite{balakrishnan2019voxelmorph}.

In this context, the emergence of foundation models \cite{tian2024unigradicon,wang2024advancing,demir2024multigradicon,zhang2024challenges} provides a new approach to address this issue. These models leverage deep learning to train registration networks on multiple different datasets, thereby obtaining a universal image registration foundation model. This model can be successfully applied to image registration tasks across different image sources, anatomical regions, and imaging modalities.

Although foundation models have shown great potential in the field of medical image registration, their application is still in the early stages, particularly in improving generalization and robustness. Many deep learning-based registration models rely on specific datasets during training, which limits their performance when handling new anatomical structures, different imaging conditions, or unseen modalities. For instance, when the model processes images from different hospitals or devices, its registration accuracy often declines significantly.
Furthermore, these models typically require large amounts of high-quality, accurately annotated training data to achieve good performance. However, annotating medical image data is costly, and due to ethical and privacy concerns, obtaining such data often faces many challenges. Therefore, enhancing the generalization ability and robustness of foundation models remains a major challenge in current research.

This study proposes introducing Sharpness-Aware Minimization (SAM) \cite{foret2020sharpness,lin2024robust,lin2024robust1,lin2024preserving} to further enhance the generalization ability of foundation models in medical image registration. SAM optimizes the flattening of the loss surface, improving the model's stability across different data distributions, which helps tackle the diversity and complexity encountered in real-world clinical environments. This study aims to validate the application value of SAM in foundation models for medical image registration and explore its potential in enhancing the model’s cross-dataset generalization ability.

In summary, our work makes the following contributions:

\begin{itemize}
    \item This study is the first to introduce SAM into foundation models for medical image registration, with the aim of improving the generalizability of the model in different data distributions by optimizing the flatness of the loss surface.
    \item Our experiments on multiple different datasets show that the proposed method outperforms other state-of-the-art methods, providing a promising approach for foundation model research.
\end{itemize}

\section{Related work}
\subsection{Traditional Medical Image Registration Methods}
Traditional medical image registration methods primarily rely on optimization algorithms to establish spatial transformation relationships by maximizing the similarity measures between images. Some studies perform optimization in the displacement vector field space, including elastic models \cite{bajcsy1989multiresolution,davatzikos1997spatial}, B-spline-based free-form deformations \cite{rueckert1999nonrigid}, discrete methods \cite{dalca2016patch,glocker2008dense}, and the Demons algorithm \cite{thirion1998image}. Topology-preserving diffeomorphic transformations have shown significant advantages in various registration tasks, including large diffeomorphic distance metric mapping (LDDMM) \cite{beg2005computing,zhang2017frequency}, diffeomorphic Demons \cite{vercauteren2009diffeomorphic}, and standard symmetric normalization (Syn) \cite{avants2008symmetric}. However, despite their solid theoretical foundation and high interpretability, all of these non-learning-based methods optimize an energy function for each image pair, which leads to limitations in computational efficiency. In particular, when handling large-scale and complex medical images, these methods often require substantial computation time and resource consumption \cite{modat2010fast,modat2014global}.

\subsection{Deep Learning-Based Medical Image Registration Methods}
With the introduction of VoxelMorph \cite{balakrishnan2019voxelmorph}, the first end-to-end unsupervised learning medical image registration network based on convolutional neural networks (CNN) \cite{gu2018recent}, the use of neural networks to learn the function of medical image registration has become a trend for the future. This includes networks like the Deep Laplacian Pyramid Image Registration Network (LapIRN) \cite{mok2020large}, the CycleMorph \cite{kim2021cyclemorph} based on Generative Adversarial Networks (GANs) \cite{goodfellow2020generative}, and the DiffuseMorph \cite{kim2022diffusemorph} based on diffusion models \cite{ho2020denoising}. Recently, the Vision Transformer architecture has been proposed. Due to the significant increase in receptive field size provided by Transformer \cite{vaswani2017attention}, it can more accurately understand the spatial correspondence between the moving and fixed images, making it a strong candidate for image registration. Therefore, the first hybrid Transformer-CNN model for volumetric medical image registration, TransMorph \cite{chen2022transmorph}, was proposed. Both qualitative and quantitative results show that Transformer-based models significantly outperform CNN-based methods in terms of performance, proving the effectiveness and advantages of Transformer in medical image registration. Compared to traditional methods, deep learning-based image registration methods can significantly improve the speed and accuracy of registration and handle more complex nonlinear transformations. However, these methods are typically optimized for specific datasets and tasks, and therefore remain significantly less flexible than traditional methods that rely on numerical optimization for each image pair.

\subsection{Foundation Models in Medical Image Registration}
Foundation Models \cite{tian2024unigradicon,wang2024advancing,demir2024multigradicon,zhang2024challenges}, as a new research direction, have gained widespread attention in the field of medical imaging in recent years. These models learn general image registration features and transformation patterns through pretraining on large and diverse datasets, enabling them to be transferred and applied to various tasks and datasets, such as the uniGradICON model. Trained on large-scale datasets, these models have shown strong performance and can effectively register images across different types of medical images and anatomical regions. However, despite the promising potential of these models, they still face several challenges, particularly in terms of model generalization and limitations in cross-domain applications.
\subsection{SAM Method and Its Applications in Image Processing}
Sharpness-Aware Minimization (SAM) \cite{foret2020sharpness} is an optimization method that has gained attention in recent years, aiming to enhance model robustness and generalization ability by adjusting the flatness of the loss surface. The core idea of SAM is to improve the model's stability and generalization ability across different data distributions by minimizing the "sharpness" of the loss during the training process. This method has effectively improved model performance in a variety of tasks, particularly in areas such as image classification and natural language processing, where SAM has demonstrated significant advantages in boosting model generalization. However, the application of SAM in the field of medical image registration is still in its early stages, and its potential for enhancing foundational models has yet to be fully explored, undoubtedly providing researchers in this field with new ideas and directions.

\section{Method}
\subsection{Registration Network}

\begin{figure}
    \centering
    \includegraphics[width=1\linewidth]{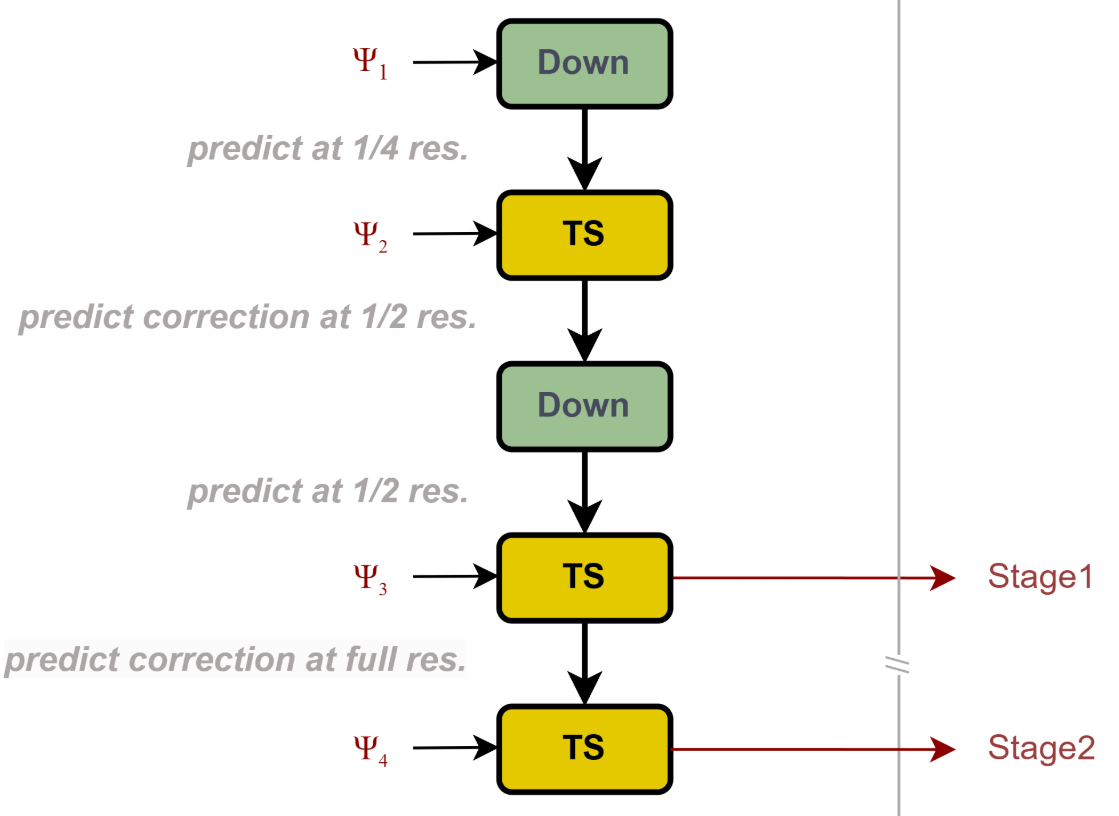}
    \caption{The illustration shows the combination steps to construct our registration network through downsampling (Down) and the two-step (TS) operator.}
    \label{fig:Architecture Diagram}
\end{figure}

% We use the publicly available GradICON medical image registration network[26]. It employs a two-step registration process:

% 1) the input images are processed through a three-level multi-resolution registration network, with each level using a UNet[23] structure that receives the deformed image from the previous level and the target image. The input images are downsampled at different levels to resolutions of 1/4, 1/2, and 1 to optimize the processing.

% 2) The image is then processed through another UNet, which receives the deformed image and target image from the first step, using full-resolution inputs. All four UNet architectures are the same.
To succinctly describe our network structure, we represent a registration neural network as $\Phi$(or e.g. $\Psi$) \cite{tian2023gradicon}. The notation $\Phi^{AB}$ (shorthand for $\Phi [ I^{A},I^{B} ] $) represents the output of this network (a transform from $\mathbb{R}^{d} \rightarrow \mathbb{R}^{d}$) for input images $I^{A}$ and $I^{B}$. 
To combine such registration networks into a multistep, multiscale approach,
we rely on the following combination operators from ICON \cite{greer2021icon}:

% \makeatletter
% \renewcommand{\maketag@@@}[1]{\hbox{\m@th\normalsize\normalfont#1}}%
% \makeatother
\begin{footnotesize}
\begin{align}
    Down\left\{\Phi \right\}\left[I^{A} , I^{B}\right]&=\Phi \left[AvgPool\left ( I^{A},2\right),AvgPool\left(I^{B},2\right)\right]\\
    TS\left \{ \Phi, \Psi  \right \} \left [ I^{A},I^{B}  \right ] &=\Phi \left [  I^{A},I^{B}\right ]\circ\Psi \left [ I^{A} \circ\Phi \left [  I^{A},I^{B}\right ],I^{B}\right ]
\end{align}
\end{footnotesize}
The downsample operator (Down) predicts the warp between two high-resolution images using a network that operates on low-resolution images, while the two-step operator (TS) predicts the warp between two images in two steps, first capturing the coarse transform via $\Phi$, and then the residual transform via $\Psi$. We combine these operators to realize a multi-resolution, multi-step network.see Fig. \ref{fig:Architecture Diagram}, via
\makeatletter
\renewcommand{\maketag@@@}[1]{\hbox{\m@th\normalsize\normalfont#1}}%
\makeatother
\begin{align}
    Stage1&=TS\left \{ Down \left \{ TS\left \{ Down \left \{ \Psi_{1}  \right \}, \Psi_{2}\right \} \right \} , \Psi_{3}\right \} \\
    Stage2&=TS\left \{Stage1,\Psi_{4}  \right \} 
\end{align}
Our atomic registration networks $\Psi_{i}$ are each represented by a UNet \cite{ronneberger2015u} instance from ICON, where each network takes two images as input and outputs a displacement field $D$. These displacement fields are then converted into functions $x\mapsto x+interpolate(D,x)$, because the above operators are defined on networks that return functions from $ \mathbb{R} ^{d}$ to $\mathbb{R} ^{d}$.
\subsection{Sharpness-Aware Minimization (SAM)}
We provide a brief introduction to the SAM algorithm. Interested readers can refer to the original paper for a detailed explanation \cite{foret2020sharpness}. In our introduction, we use the $\ell$2 norm(p = 2 using notation from the original paper), Suppose the Adam optimizer is used, and an approximate method proposed by Brock et al. is employed to efficiently compute the adversarial points \cite{brock2021high}. Given a loss function $L:\mathcal{W}\times\mathcal{X}\times\mathcal{Y}\to\mathbb{R}_+$ , SAM aims to find the parameter \(\omega\) whose neighborhood has low training loss by optimizing the minimax objective:

\begin{equation}
\underset{\omega}{min} \underset{\left \| \epsilon  \right \|_{2} \le \rho  }{max} L_{train} \left ( \omega  + \epsilon  \right).
\end{equation}
Finding the exact optima \(\epsilon ^{\ast }  \) of the inner maximization is challenging, so we employ a first-order approximation, resulting in:
\begin{align}
\hat{\epsilon } \left ( \omega  \right ) &= \underset{\left \| \epsilon  \right \| _{2}\le \rho  }{argmin}L_{train}\left ( \omega  \right ) +  \epsilon ^{T}\bigtriangledown _{\omega L_{train}\left ( \omega  \right )  }\\
&= \rho \bigtriangledown _{\omega } L_{train} \left ( \omega  \right ) /\left \| \bigtriangledown _{\omega } L_{train}\left ( \omega  \right )   \right \| _{2}.
\end{align}
That is, \( \hat{\epsilon} \) is just a scaling of the loss gradient at the current parameters. After computing \( \hat{\epsilon} \left ( \omega  \right ) \), SAM performs gradient descent using the gradient \(\bigtriangledown _{\omega } L_{train}\left ( \omega  \right ) | _{\omega _{adv} }  \) at the nearly "adversarial" point \(\omega _{adv}\left ( \omega  \right )\overset{\bigtriangleup}{=}\omega +  \hat{\epsilon }  \left ( \omega  \right ) \).

\begin{figure}[t]
    \centering
    \setlength{\fboxsep}{0pt} 
    \fbox{\includegraphics[width=0.75\linewidth]{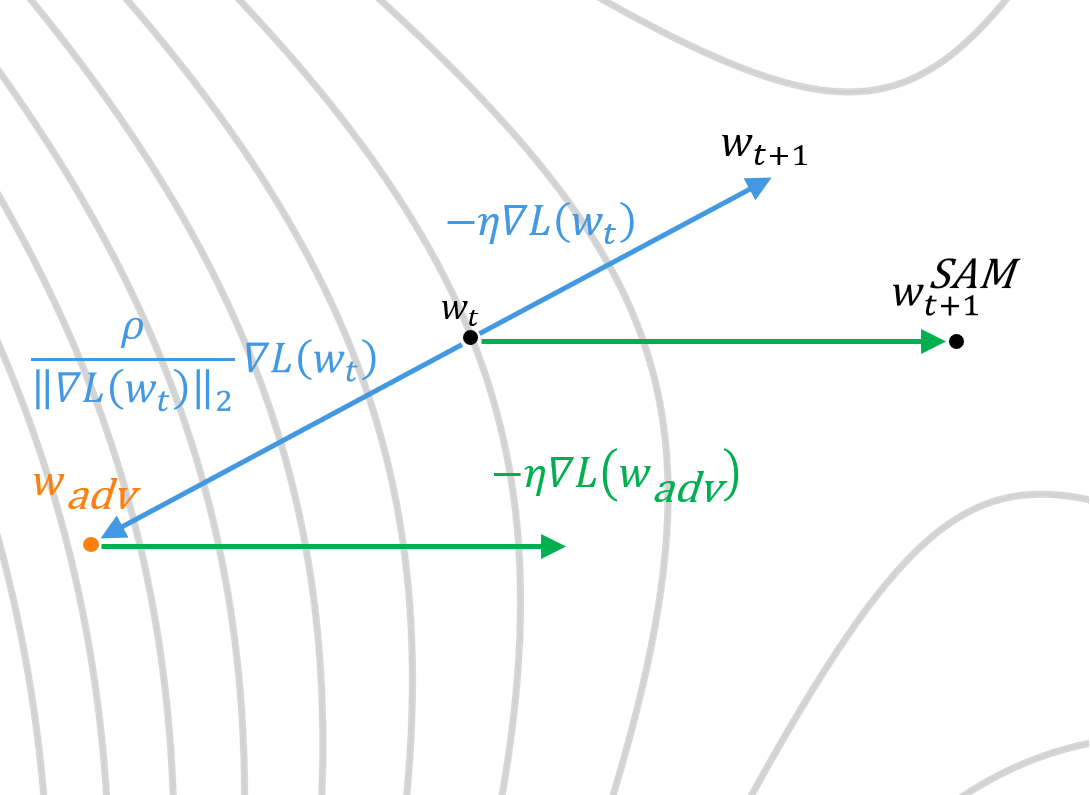}}
    \caption{Schematic of the SAM parameter update.}
    \label{fig:Diagram of SAM}
\end{figure}

Based on the principles of SAM, we can deduce that SAM is compatible with any first-order optimizer by simply replacing the gradient of the mini-batch B at the current model weights \(\omega _{adv}\in W \) with the gradient computed at \(\omega _{adv} \) \cite{}. Algorithm \ref{alg:SAM} gives pseudo-code for the full SAM algorithm, using Adam as the base optimizer, and Fig. \ref{fig:Diagram of SAM}  schematically illustrates a single SAM parameter update \cite{bahri2021sharpness}.

\begin{algorithm}[t]
    \caption{SAM Algorithm.}
    \label{alg:SAM}
    \renewcommand{\algorithmicrequire}{\textbf{Input:}}
    \renewcommand{\algorithmicensure}{\textbf{Output:}}
    \begin{algorithmic}
        \REQUIRE Training set ${\mathcal{S}} \overset {\triangle}{=}\mathrm {U}_{i=1}^{n} \left\{ \left ( x_{i} ,y_{i}\right )\right\} $ ,loss function $L:\mathcal{W} \times \mathcal{X} \times \mathcal{Y} \longrightarrow \mathbb{R_{+}}$ ,batch size b,step size $\eta > 0$, neighborhood size $\rho > 0$ (default 0.15)%%input
        \ENSURE Model trained with SAM   %%output
        \STATE  Initialize weights $w_{0},t=0$;
        \WHILE{$not$ $converged$}
            \STATE Sample batch $\mathcal{B}=\left\{\left(x_{1},y_{1},...(x_{b},y_{b})\right)\right\}$;\STATE Compute gradient$\bigtriangledown_{w}L_{\mathcal{B}}\left(w\right) $ of the batch's training loss;\STATE Compute $\hat{\epsilon}\left(w\right)$ per equation 7;
            \STATE Compute gradient approximation for the SAM objective:\STATE$g=\bigtriangledown_{w}L_{\mathcal{B}}\left(w\right)|_{w+\hat{\epsilon}(w)}$;
            \STATE Update weights: $w_{t+1}=w_{t}-\eta g$;
            \STATE t=t+1;
        \ENDWHILE
        \RETURN $w_{t}$
    \end{algorithmic}
\end{algorithm}

\textbf{Taining loss.} The loss function proposed in GradICON \cite{tian2023gradicon} has the following form:
\makeatletter
\renewcommand{\maketag@@@}[1]{\hbox{\m@th\normalsize\normalfont#1}}%
\makeatother
\begin{align} \label{eq:8}
    L &= L_{sim}\left(I^{A}\circ\Phi^{AB},I^{B}\right) \notag \\
      &\quad + L_{sim}\left(I^{B}\circ\Phi^{BA},I^{A}\right) \notag \\
      &\quad + \lambda\left\| \nabla (\Phi^{AB}\circ\Phi^{BA}) - I \right\|_{F}^{2}
\end{align}
Given an ordered image pair ($I^{A}$,$I^{B}$), the registration network outputs the transformation map $\Phi^{AB}$ which maps $I^{A}$ to the space of $I^{B}$. By swapping the input pair ($I^{B}$,$I^{A}$), we obtain the estimated inverse map $\Phi^{BA}$. The loss of similarity $L_{sim}$ is calculated between the warped image \(I^{A}\circ\Phi^{AB}\)
and the target image $I^{B}$, and vice versa. We use localized normalized cross correlation $(1 - LNCC) $ as similarity measure. The third term in Eq. (\ref{eq:8}) is the gradient inverse consistency
regularizer, which penalizes differences between the Jacobian of the composition of $\Phi^{AB}$ with $\Phi^{BA}$ and the indentity matrix $I$; $\left \| \cdot  \right \| _{F}^{2} $ is the Frobenius norm, $\lambda >  0$.

The reason for choosing GradICON as the basic building block of the foundational model is that it can maintain consistent hyperparameters and training settings across various datasets while demonstrating excellent registration performance \cite{tian2023gradicon}. This characteristic ensures that we can train a robust foundational registration model on the composite dataset, offering an advantage over methods that rely on task-specific hyperparameters and training settings.

\section{Experiment}

\subsection{Training Dataset and Evaluation Metric.} 

\textbf{Dataset.} Our composite dataset (see Table \ref{tab:t1}) contains both intra-patient and inter-patient data. The intra-patient dataset (Dataset 1) consists of 899 pairs of inspiratory/expiratory lung CT images. The inter-patient dataset contains 2532, 1076, and 30 images. We randomly selected two images from each dataset, resulting in 3,205,512, 578,888, and 450 possible different image pairs for Datasets 2-4. To avoid bias caused by different numbers of paired images across datasets, we randomly sampled N = 1000 image pairs from each dataset during each training cycle, yielding a total of 4000 3D image pairs per cycle. These are the default settings for uniGradICON \cite{tian2024unigradicon}. All images were resampled to [128, 128, 128] and then normalized to [0, 1].

\begin{table*}
    \centering
    \caption{Training and Testing Dataset}
    \label{tab:t1}
    \renewcommand{\arraystretch}{1.25}
    \begin{tabular}{ccccccc}
        \hline
         Dataset & Anatom.region & \# of patients & \# per patients & \# of pairs & Type & Modality \\ 
        1.COPDGene \cite{regan2011genetic} & Lung & 899 & 2 & 899 & Intra-pat. & CT  \\ 
        2.OAI \cite{nevitt2006osteoarthritis} & Knee & 2532 & 1 & 3,205,512 & Inter-pat. & MRI  \\ 
        3.HCP \cite{van2012human} & Brain & 1076 & 1 & 578,888 & Inter-pat. & MRI  \\ 
        4.L2R-Abdomen \cite{xu2016evaluation} & Abdomen & 30 & 1 & 450 & Inter-pat. & CT  \\ \hline
        5.ACDC-test \cite{bernard2018deep} & Cardiac & 50 & 2 & 50 & Intra-pat. & MRI  \\ 
        6.SLIVER \cite{heimann2009comparison} & Liver & 20 & 1 & 19 & Inter-pat. & MRI  \\ 
        7.L2R-AbdomenMRCT-test \cite{hering2022learn2reg} & Abdomen & 8 & 1 & 8 & Intra-pat. & MRI/CT  \\ 
        8.L2R-HippocampusMR-test \cite{hering2022learn2reg} & Hippocampus & 131 & 1 & 130 & Inter-pat. & MRI  \\ 
        9.L2R-LungCT-test \cite{hering2022learn2reg} & Lung & 10 & 1 & 9 & Inter-pat. & CT \\ \hline
    \end{tabular}
\end{table*}

\textbf{Evaluation Metrics.} We evaluated the registration performance using the Dice coefficient and the Jacobian determinant. The registration performance of each model was evaluated based on the volumetric overlap of anatomical structures/organs, quantified by the Dice score. The Dice scores for all anatomical structures / organs in all patients were averaged. Subsequently, the mean and standard deviation of Dice scores were compared between different registration methods.

The Jacobian matrix $ J_{\phi } \left ( v \right ) =\bigtriangledown \phi \left ( v \right ) \in R^{3\times 3}$ was calculated to capture the local properties around the voxel $v$ . Voxels with $J_{\phi } \le 0$ were recorded as singular points, indicating folding in the image. We calculated the proportion of $J_{\phi } \le 0 $ [\%] for each deep learning-based method to quantitatively measure the texture-preserving capability of the deformation field. A smaller proportion indicates better texture preservation in the deformation \cite{chen2023transmatch}.

\subsection{Implementation details}

\textbf{Training hyperparameters.} The proposed methodology was implemented using the PyTorch framework, and training and evaluation were performed on a single NVIDIA GeForce RTX 3090 GPU. We trained for 800 epochs in the first step and 200 epochs in the second step, with a learning rate of $\eta = 5e-5 $  and a balancing constant $\lambda  = 1.5$. These are the default settings for GradICON \cite{tian2023gradicon}. SAM has a single hyperparameter $\rho$, which is the size of the step taken along the unit adversarial gradient vector and was set to 0.1.

\subsection{Comparisons With the State-of-the-Art Approaches}
\textbf{Baseline Comparisons.} To validate the performance improvement of the proposed method, we adopted several comparison methods that have demonstrated outstanding performance in image registration: Elastix \cite{klein2009elastix}, VoxelMorph \cite{balakrishnan2019voxelmorph}, DiffuseMorph \cite{kim2022diffusemorph}, TransMorph \cite{chen2022transmorph}, and uniGradICON \cite{tian2024unigradicon}.

To evaluate the performance of the proposed method in cardiac MR image registration, we compared it with several baseline methods: the traditional method Elastix, the convolutional neural network-based method VoxelMorph, the diffusion model-based method DiffuseMorph, and the Transformer-based method TransMorph. Table {\ref{tab:tab2}} shows the quantitative evaluation results of the average Dice score and the percentage of non-positive Jacobian determinant values across all structures and scans for the cardiac MRI data (i.e., the ACDC dataset) and liver MRI data (i.e., the SLIVER dataset). A higher Dice score indicates better overlap of the anatomical structures after registration, while a lower percentage of voxels with non-positive Jacobian determinant values suggests smoother and more reasonable deformation, avoiding voxel folding. For cardiac MRI image registration, our proposed method achieved the highest average Dice score of 0.7360, ranking first among all methods. At the same time, it attained a non-positive Jacobian determinant percentage of only 6.29e-7, which is significantly lower than that of other methods. This indicates that our method not only ensures registration accuracy but also provides a more stable and smoother deformation field. For liver MRI image registration, our method also achieved the highest average Dice score of 0.8722 and was the only model to reach nonpositive Jacobian determinant values of 0. This means that there were no voxel foldings during registration, which demonstrated excellent regularity and stability of the deformation. In particular, compared to traditional medical image registration methods such as Elastix, our method achieved approximately a 3\% improvement in Dice score on both datasets and significantly reduced the occurrence of folding regions, fully demonstrating its superior balance between registration accuracy and diffeomorphic properties.

\begin{table}
    \centering
    \caption{The quantitative evaluation results of cardiac MRI data (i.e., the ACDC dataset) and liver MRI data (i.e., the SLIVER dataset). Dice score and percentage of voxels with a non-positive Jacobian determinant (i.e., folded voxels) are evaluated for different methods. The bolded numbers denote the highest scores, while the italicized ones indicate the second highest.} 
    \label{tab:tab2}
    \renewcommand{\arraystretch}{1.25}
    \begin{tabular}{ccccc}
        \hline
         \multirow{2}{*}{methods}& \multicolumn{2}{c}{ACDC} & \multicolumn{2}{c}{SLIVER}\\
         & Dice & $\%\left | J \right | _{< 0} $ & Dice & $\%\left | J \right | _{< 0} $\\
         \hline
         Initial& 0.54 & 0 & 0.7546 & 0 \\
         Elastix & 0.7032 & \textit{1.186e-5} & 0.8407(0.0324) &1.836e-2 \\
         VoxelMorph  & 0.6230 & 2.459e-3 & \textit{0.8680(0.0362)} & 5.469e-2\\
         DiffuseMorph & 0.6951 & 2.074e-4 & 0.8071(0.0588) & \textit{1.683e-2}\\
         TransMorph & 	\textit{0.7322}&4.254e-3&0.8646(0.0343)&4.130e-1 \\
         Ours&\textbf{0.7360}&\textbf{6.29e-7}&\textbf{0.8722}(0.0398)&	\textbf{0} \\
         \hline
    \end{tabular}
\end{table}

\begin{figure}
    \centering
    \includegraphics[width=1\linewidth]{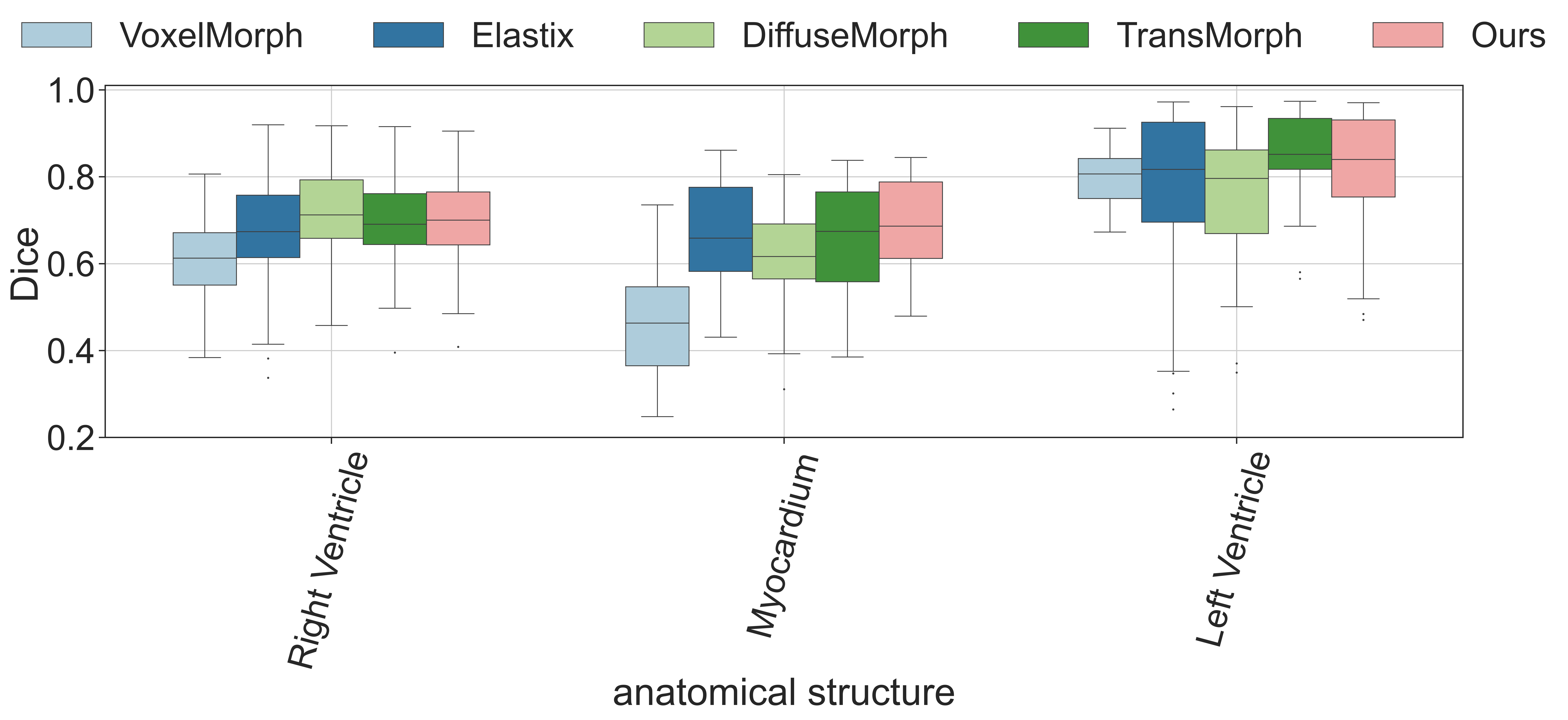}
    \caption{The boxplot shows the Dice scores of Voxelmorph, Elastix, DiffuseMorph, TransMorph, and our proposed method on 3 anatomical structures in the ACDC dataset.}
    \label{fig:fig3}
\end{figure}

As shown in Fig. {\ref{fig:fig3}}, the boxplot presents the Dice scores for three different anatomical structures in the ACDC test scans. Although our method achieves slightly lower Dice scores for the left ventricle compared to TransMorph and for the right ventricle compared to DiffuseMorph, it still attains the second-highest Dice scores in both regions and achieves the highest Dice score for the myocardium. Overall, our method achieves the highest average Dice score across the three anatomical structures.

% \begin{figure}
%     \centering
%     \includegraphics[width=1\linewidth]{final_combined_image.png}
%     \caption{An example of axial slices for cardiac MR image registration, displaying the results of all comparison methods. The rows are the target fixed image, the transformed moved image after registration (moved),the registration error map (white indicates zero error, red indicates positive error, blue indicates negative error), 
%     the registration transformation, the registered label map (green represents the right ventricle, dark blue represents the myocardium, and light blue represents the left ventricle)}
%     \label{fig:enter-label}
% \end{figure}
% \begin{figure}
%     \centering
%     \includegraphics[width=1\linewidth]{final_combined_image(1)(1).png}
%     \caption{An example of axial slices for cardiac MR image registration, displaying the results of all comparison methods. The rows are the target fixed image, the transformed moved image after registration (moved),the registration error map (white indicates zero error, red indicates positive error, blue indicates negative error), 
%      the registration transformation, the registered label map (green represents the right ventricle, dark blue represents the myocardium, and light blue represents the left ventricle)}
%     \label{fig:enter-label}
% \end{figure}

\begin{figure}
    \centering
    \includegraphics[width=1\linewidth]{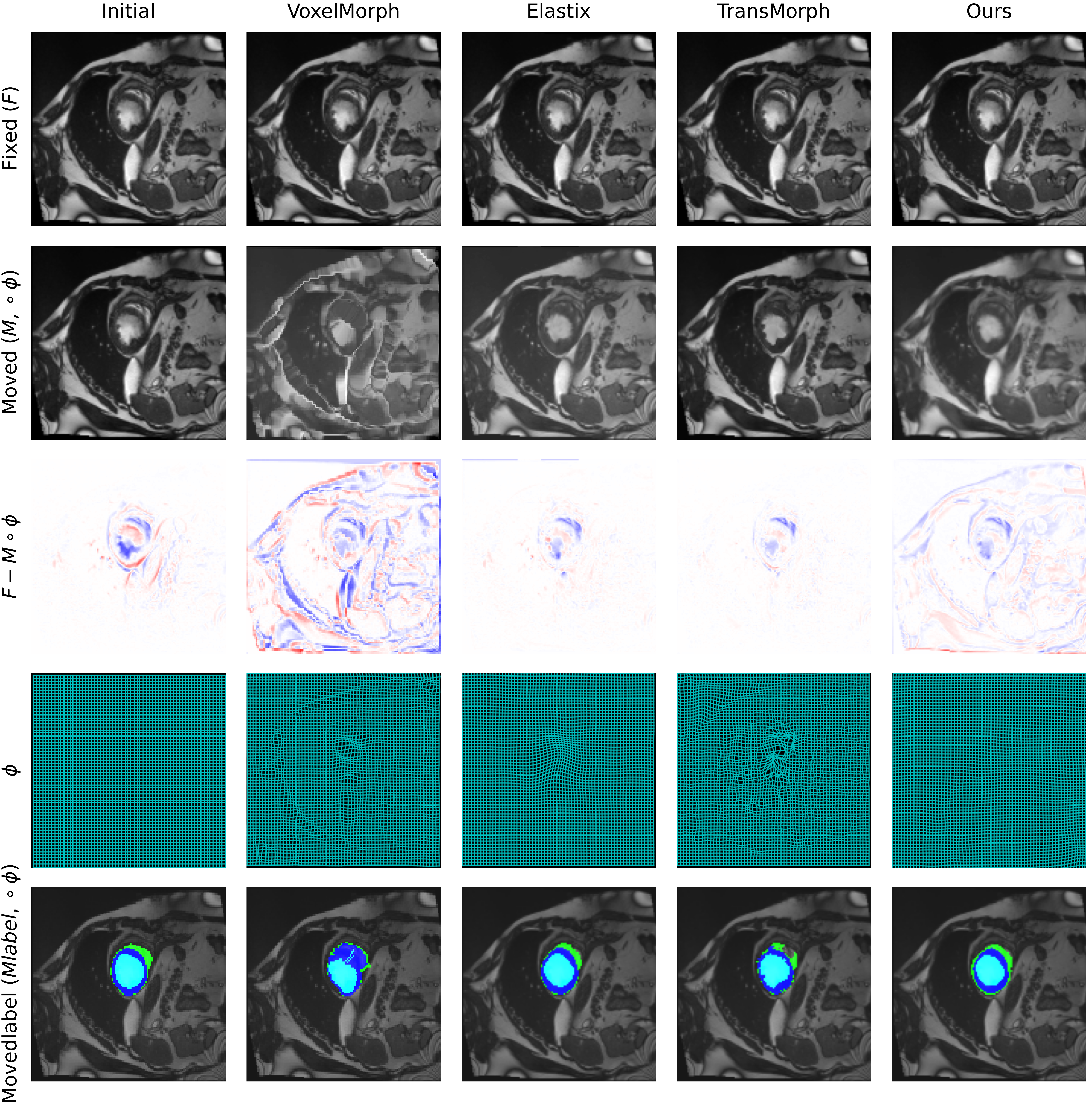}
    \vspace{-8mm}
    \caption{An example of axial slices for cardiac MR image registration, displaying the results of all comparison methods. The rows are the target fixed image, the transformed moved image after registration (moved),the registration error map (white indicates zero error, red indicates positive error, blue indicates negative error), 
     the registration transformation, the registered label map (green represents the right ventricle, dark blue represents the myocardium, and light blue represents the left ventricle)}
    \label{fig:fig4}
\vspace{-2mm}
\end{figure}

The panel of Fig. {\ref{fig:fig4}} presents the qualitative results of cardiac MRI registration on a specific slice from one patient. Based on the second row, which shows the transformed moving images after registration, qualitative analysis reveals that our method handles certain details—especially in the right ventricle region of the heart—significantly better than other methods and is closer to the fixed image. In contrast, VoxelMorph fails to clearly delineate the contour of the right ventricle.
From the third row, which displays the error maps between the registered and fixed images, although our method performs slightly worse than Elastix and TransMorph overall, the difference map in the central region of the heart (the middle part of the image) appears significantly lighter compared to the original error map, indicating an improvement in registration accuracy. Furthermore, as shown in the fourth row, which visualizes the deformation field as a grid, our method generates a smoother displacement field than both VoxelMorph and TransMorph.
Notably, in the final row showing the registered label maps, VoxelMorph exhibits clear errors in the myocardial region, highlighting the challenge of achieving perfect registration across all anatomical structures. Although TransMorph generally provides accurate label registration, some discontinuous and missing areas are observed. In contrast, our method, along with Elastix, performs closer to the ground truth in the registered label map, and our method yields higher Dice coefficients for the segmented structures.

\subsection{Ablation Study}
\begin{figure}
    \centering
    \includegraphics[width=1\linewidth]{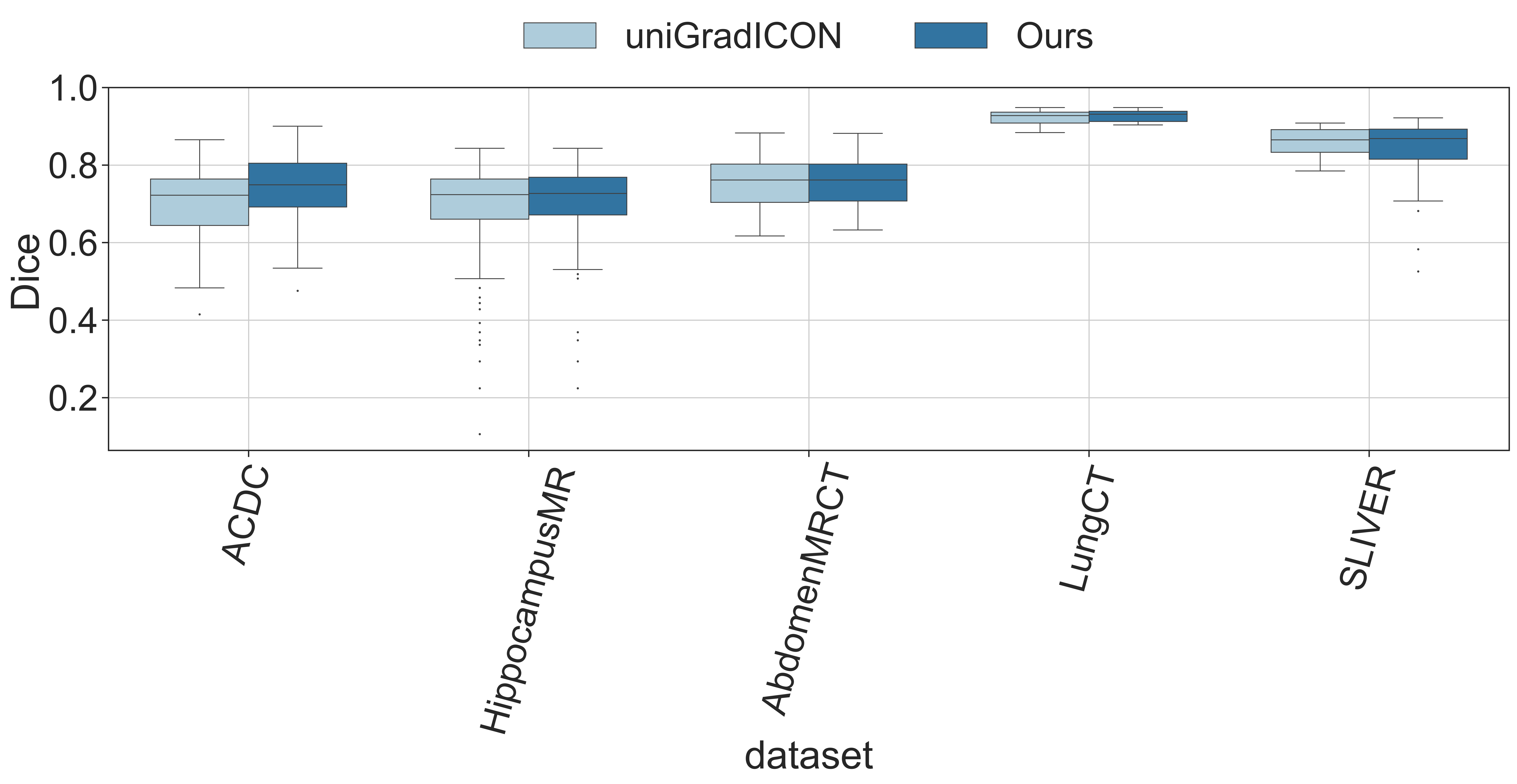}
    % \vspace{mm}
    \caption{The boxplot shows the Dice score results of our method and uniGradICON on the ACDC, HippocampusMR, AbdomenMRCT, LungCT, and SLIVER datasets.}
    \label{fig:fig5}
    % \vspace{-6mm}
\end{figure}

In addition to the cardiac MRI data (i.e., the ACDC dataset) and liver MRI data (i.e., the SLIVER dataset), we also selected test datasets from different anatomical regions provided by the Learn2Reg challenge, including abdominal MR-CT data (i.e., AbdomenMRCT), hippocampus MR data (i.e., HippocampusMR), and lung CT data (i.e., LungCT). As shown in Fig. \ref{fig:fig5}, our method is compared with uniGradICON across these five diverse datasets. It can be observed that our method consistently achieves significantly higher Dice coefficients on all five datasets, indicating that our method outperforms uniGradICON in terms of registration performance and demonstrates stronger generalization capability.

\begin{figure}[t]
    \centering
    \includegraphics[width=1\linewidth]{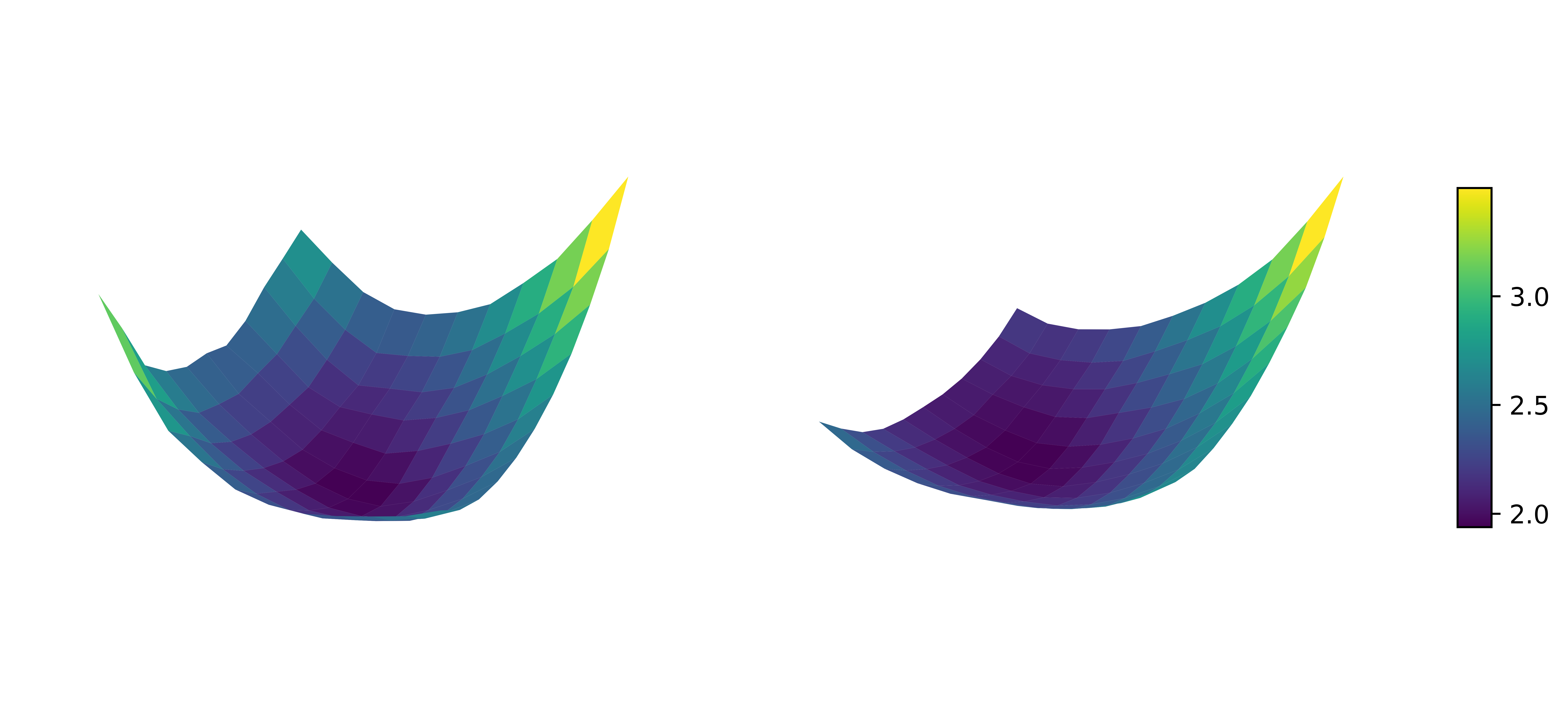}
    \vspace{-8mm}
    \caption{Loss landscape visualization. Left: uniGradICON. Right: Ours.}
    \vspace{-2mm}
    \label{fig:fig6}
    \vspace{-3mm}
\end{figure}

The geometry of the loss landscape, particularly the flatness around the minima, is directly related to a model's generalization ability. In general, a flatter loss landscape indicates better generalization. As shown in Fig. \ref{fig:fig6}, from the loss landscape visualizations of uniGradICON and our model, it can be observed that on the left side of the landscape, uniGradICON exhibits a steeper geometry compared to our model. Overall, the loss landscape of our model is flatter than that of uniGradICON, suggesting stronger generalization capability. This further supports the effectiveness of incorporating SAM into the base model.

\section{conclusion}

In this study, we proposed a general medical image registration method that integrates Sharpness-Aware Minimization (SAM) into the base model, aiming to enhance the model’s generalization capability across different anatomical regions and further improve registration accuracy. Experimental results demonstrated that the proposed method achieved significant improvements in general registration performance. By incorporating SAM into the base model, the training process can more effectively explore flatter regions of the loss landscape, thereby enhancing the model’s robustness and generalization ability. We validated these performance gains through ablation experiments.
With these enhancements, our method achieved superior registration results on the ACDC, SLIVER, AbdomenMRCT, HippocampusMR, and LungCT datasets. Compared with other models, our method attained significantly higher Dice coefficient scores and a notably lower proportion of voxels with non-positive Jacobian determinant values. Moreover, our method maintained stable performance under different imaging modalities and scanning conditions, further validating its effectiveness and great potential in general medical image registration tasks.
These findings highlight the broad application prospects of SAM in general medical image registration. Future work may explore integrating the base model with other advanced deep learning techniques to further optimize medical image registration methods.

\section*{Acknowledgment}
This work was supported in part by the National Natural Science Foundation of China under Grant 42375148, Sichuan province Key Technology Research and Development project under Grant No.2024ZHCG0190, No. 2024ZHCG0176.

\small
\bibliographystyle{IEEEtran}
\bibliography{IEEEabrv,reference.bib}

\begin{thebibliography}{10}
\providecommand{\url}[1]{#1}
\csname url@samestyle\endcsname
\providecommand{\newblock}{\relax}
\providecommand{\bibinfo}[2]{#2}
\providecommand{\BIBentrySTDinterwordspacing}{\spaceskip=0pt\relax}
\providecommand{\BIBentryALTinterwordstretchfactor}{4}
\providecommand{\BIBentryALTinterwordspacing}{\spaceskip=\fontdimen2\font plus
\BIBentryALTinterwordstretchfactor\fontdimen3\font minus \fontdimen4\font\relax}
\providecommand{\BIBforeignlanguage}[2]{{%
\expandafter\ifx\csname l@#1\endcsname\relax
\typeout{** WARNING: IEEEtran.bst: No hyphenation pattern has been}%
\typeout{** loaded for the language `#1'. Using the pattern for}%
\typeout{** the default language instead.}%
\else
\language=\csname l@#1\endcsname
\fi
#2}}
\providecommand{\BIBdecl}{\relax}
\BIBdecl

\bibitem{ashburner2007fast}
J.~Ashburner, ``A fast diffeomorphic image registration algorithm,'' \emph{Neuroimage}, vol.~38, no.~1, pp. 95--113, 2007.

\bibitem{dalca2016patch}
A.~V. Dalca, A.~Bobu, N.~S. Rost, and P.~Golland, ``Patch-based discrete registration of clinical brain images,'' in \emph{Patch-Based Techniques in Medical Imaging}.\hskip 1em plus 0.5em minus 0.4em\relax Springer, 2016.

\bibitem{glocker2008dense}
B.~Glocker \emph{et~al.}, ``Dense image registration through mrfs and efficient linear programming,'' \emph{MIA}, 2008.

\bibitem{thirion1998image}
J.-P. Thirion, ``Image matching as a diffusion process: an analogy with maxwell's demons,'' \emph{MIA}, 1998.

\bibitem{zitova2003image}
B.~Zitova and J.~Flusser, ``Image registration methods: a survey,'' \emph{Image and vision computing}, 2003.

\bibitem{maintz1998survey}
J.~A. Maintz and M.~A. Viergever, ``A survey of medical image registration,'' \emph{Medical image analysis}, 1998.

\bibitem{hu2023attention}
J.~Hu, Z.~Shuai, X.~Wang, S.~Hu, S.~Sun, S.~Lyu, and X.~Wu, ``Attention guided policy optimization for 3d medical image registration,'' \emph{IEEE Access}, vol.~11, pp. 65\,546--65\,558, 2023.

\bibitem{luo2022stochastic}
Z.~Luo, J.~Hu, X.~Wang, S.~Hu, B.~Kong, Y.~Yin, Q.~Song, X.~Wu, and S.~Lyu, ``Stochastic planner-actor-critic for unsupervised deformable image registration,'' in \emph{AAAI}, 2022.

\bibitem{kybic2003fast}
J.~Kybic and M.~Unser, ``Fast parametric elastic image registration,'' \emph{TIP}, 2003.

\bibitem{bajcsy1989multiresolution}
R.~Bajcsy \emph{et~al.}, ``Multiresolution elastic matching,'' \emph{Computer vision, graphics, and image processing}, 1989.

\bibitem{shen2002hammer}
D.~Shen \emph{et~al.}, ``Hammer: hierarchical attribute matching mechanism for elastic registration,'' \emph{TMI}, 2002.

\bibitem{liu2011simple}
Z.~Liu \emph{et~al.}, ``A simple and robust feature point matching algorithm based on restricted spatial order constraints for aerial image registration,'' \emph{IEEE Transactions on Geoscience and Remote Sensing}, 2011.

\bibitem{thirion1996new}
J.-P. Thirion, ``New feature points based on geometric invariants for 3d image registration,'' \emph{IJCV}, 1996.

\bibitem{klein2009elastix}
S.~Klein \emph{et~al.}, ``Elastix: a toolbox for intensity-based medical image registration,'' \emph{TMI}, 2009.

\bibitem{rohde2003adaptive}
G.~K. Rohde \emph{et~al.}, ``The adaptive bases algorithm for intensity-based nonrigid image registration,'' \emph{TMI}, 2003.

\bibitem{modat2010fast}
M.~Modat \emph{et~al.}, ``Fast free-form deformation using graphics processing units,'' \emph{Computer methods and programs in biomedicine}, 2010.

\bibitem{modat2014global}
------, ``Global image registration using a symmetric block-matching approach,'' \emph{Journal of medical imaging}, 2014.

\bibitem{lecun2015deep}
Y.~LeCun, Y.~Bengio, and G.~Hinton, ``Deep learning,'' \emph{nature}, vol. 521, no. 7553, pp. 436--444, 2015.

\bibitem{goodfellow2016deep}
I.~Goodfellow, Y.~Bengio, A.~Courville, and Y.~Bengio, \emph{Deep learning}.\hskip 1em plus 0.5em minus 0.4em\relax MIT press Cambridge, 2016, vol.~1, no.~2.

\bibitem{huang2025diffusion}
P.~Huang, S.~Hu, B.~Peng, J.~Zhang, H.~Zhu, X.~Wu, and X.~Wang, ``Diffusion-empowered autoprompt medsam,'' \emph{arXiv preprint arXiv:2502.06817}, 2025.

\bibitem{wang2025graph}
Z.~Wang, Z.~Yin, Y.~Zhang, L.~Yang, T.~Zhang, N.~Pissinou, Y.~Cai, S.~Hu, Y.~Li, L.~Zhao \emph{et~al.}, ``Graph fairness via authentic counterfactuals: Tackling structural and causal challenges,'' \emph{ACM SIGKDD Explorations Newsletter}, 2025.

\bibitem{wang2025fg}
------, ``Fg-smote: Towards fair node classification with graph neural network,'' \emph{ACM SIGKDD Explorations Newsletter}, vol.~26, no.~2, pp. 99--108, 2025.

\bibitem{ren2024improving}
H.~Ren, L.~Li, C.-H. Liu, X.~Wang, and S.~Hu, ``Improving generalization for ai-synthesized voice detection,'' \emph{AAAI}, 2024.

\bibitem{lin2025ai}
L.~Lin, X.~Wang, S.~Hu \emph{et~al.}, ``Ai-face: A million-scale demographically annotated ai-generated face dataset and fairness benchmark,'' \emph{CVPR}, 2025.

\bibitem{fu2024vb}
Y.~Fu, X.~Zhu, X.~Li, X.~Wang, X.~Wu, S.~Hu, Y.~Wu, S.~Lyu, and W.~Liu, ``Vb-kgn: Variational bayesian kernel generation networks for motion image deblurring,'' \emph{TMM}, 2024.

\bibitem{chen2024self}
H.~Chen, H.~Guo, B.~Hu, S.~Hu, J.~Hu, S.~Lyu, X.~Wu, and X.~Wang, ``A self-learning multimodal approach for fake news detection,'' \emph{arXiv}, 2024.

\bibitem{wang2024u}
X.~Wang, X.~Liu, P.~Huang, P.~Huang, S.~Hu, and H.~Zhu, ``U-medsam: Uncertainty-aware medsam for medical image segmentation,'' \emph{arXiv preprint arXiv:2408.08881}, 2024.

\bibitem{sun2024repmedgraf}
R.~Sun, F.~Liao, J.~He, Y.~Liu, W.~Yi, Q.~Fan, S.~Hu, X.~Wang, and J.~Hu, ``Repmedgraf: Re-parameterization medical generated radiation field for improved 3d image reconstruction,'' in \emph{AVSS}.\hskip 1em plus 0.5em minus 0.4em\relax IEEE, 2024.

\bibitem{zhu2024cgd}
X.~Zhu, T.~Liu, Z.~Liu, O.~Shaobo, X.~Wang, S.~Hu, and F.~Ding, ``Cgd-net: A hybrid end-to-end network with gating decoding for liver tumor segmentation from ct images,'' in \emph{AVSS}.\hskip 1em plus 0.5em minus 0.4em\relax IEEE, 2024.

\bibitem{zhu2024qrnn}
X.~Zhu, Z.~Yan, W.~Wang, S.~Hu, X.~Wang, and C.~Liu, ``Qrnn-transformer: Recognizing textual entailment,'' in \emph{AVSS}.\hskip 1em plus 0.5em minus 0.4em\relax IEEE, 2024.

\bibitem{qiu2024enhancing}
M.~Qiu, W.~Lin, S.~Chien, L.~Christopher, Y.~Chen, and S.~Hu, ``Enhancing vehicle re-identification and matching for weaving analysis,'' in \emph{AVSS}.\hskip 1em plus 0.5em minus 0.4em\relax IEEE, 2024.

\bibitem{peng2024uncertainty}
Y.~Peng, H.~Chen, C.-S. Lin, G.~Huang, J.~Hu, H.~Guo, B.~Kong, S.~Hu, X.~Wu, and X.~Wang, ``Uncertainty-aware explainable recommendation with large language models,'' in \emph{IJCNN}.\hskip 1em plus 0.5em minus 0.4em\relax IEEE, 2024, pp. 1--8.

\bibitem{yang2024llm}
H.~Yang, H.~Chen, H.~Guo, Y.~Chen, C.-S. Lin, S.~Hu, J.~Hu, X.~Wu, and X.~Wang, ``Llm-medqa: Enhancing medical question answering through case studies in large language models,'' \emph{IJCNN}, 2025.

\bibitem{qiu2024real}
M.~Qiu, W.~Lin, L.~A. Christopher, S.~Chien, Y.~Chen, and S.~Hu, ``Real-time lane-wise traffic monitoring in optimal rois,'' \emph{MIPR}, 2024.

\bibitem{chen2024masked}
T.~Chen, S.~Yang, S.~Hu, Z.~Fang, Y.~Fu, X.~Wu, and X.~Wang, ``Masked conditional diffusion model for enhancing deepfake detection,'' \emph{arXiv preprint arXiv:2402.00541}, 2024.

\bibitem{lin2024detecting}
L.~Lin, N.~Gupta, Y.~Zhang, H.~Ren, C.-H. Liu, F.~Ding, X.~Wang, X.~Li, L.~Verdoliva, and S.~Hu, ``Detecting multimedia generated by large ai models: A survey,'' \emph{arXiv preprint arXiv:2402.00045}, 2024.

\bibitem{fan2024efficient}
Q.~Fan, C.~Wu, S.~Hu, X.~Wu, X.~Wang, and J.~Hu, ``Efficient image super-resolution via symmetric visual attention network,'' in \emph{IJCNN}.\hskip 1em plus 0.5em minus 0.4em\relax IEEE, 2024.

\bibitem{hu2024fairness}
S.~Hu and G.~H. Chen, ``Fairness in survival analysis with distributionally robust optimization,'' \emph{Journal of Machine Learning Research}, vol.~25, no. 246, pp. 1--85, 2024.

\bibitem{fan2024synthesizing}
B.~Fan, S.~Hu, and F.~Ding, ``Synthesizing black-box anti-forensics deepfakes with high visual quality,'' in \emph{ICASSP}.\hskip 1em plus 0.5em minus 0.4em\relax IEEE, 2024.

\bibitem{hu2024umednerf}
J.~Hu, Q.~Fan, S.~Hu, S.~Lyu, X.~Wu, and X.~Wang, ``Umednerf: Uncertainty-aware single view volumetric rendering for medical neural radiance fields,'' in \emph{ISBI}.\hskip 1em plus 0.5em minus 0.4em\relax IEEE, 2024, pp. 1--4.

\bibitem{zhang2024x}
L.~Zhang, H.~Chen, S.~Hu, B.~Zhu, C.-S. Lin, X.~Wu, J.~Hu, and X.~Wang, ``X-transfer: A transfer learning-based framework for gan-generated fake image detection,'' in \emph{IJCNN}.\hskip 1em plus 0.5em minus 0.4em\relax IEEE, 2024.

\bibitem{xiang2023rmbench}
Y.~Xiang, X.~Wang, S.~Hu, B.~Zhu, X.~Huang, X.~Wu, and S.~Lyu, ``Rmbench: Benchmarking deep reinforcement learning for robotic manipulator control,'' in \emph{IROS}.\hskip 1em plus 0.5em minus 0.4em\relax IEEE, 2023.

\bibitem{yang2023improving}
S.~Yang, S.~Hu, B.~Zhu, Y.~Fu, S.~Lyu, X.~Wu, and X.~Wang, ``Improving cross-dataset deepfake detection with deep information decomposition,'' \emph{arXiv}, 2023.

\bibitem{wang2023deep}
X.~Wang, Z.~Luo, J.~Hu, C.~Feng, S.~Hu, B.~Zhu, X.~Wu, and S.~Lyu, ``Deep reinforcement learning for image-to-image translation,'' \emph{arXiv preprint arXiv:2309.13672}, 2023.

\bibitem{hu2024outlier}
S.~Hu, Z.~Yang, X.~Wang, Y.~Ying, and S.~Lyu, ``Outlier robust adversarial training,'' in \emph{Asian Conference on Machine Learning}.\hskip 1em plus 0.5em minus 0.4em\relax PMLR, 2024, pp. 454--469.

\bibitem{fan2023attacking}
B.~Fan, Z.~Jiang, S.~Hu, and F.~Ding, ``Attacking identity semantics in deepfakes via deep feature fusion,'' in \emph{MIPR}.\hskip 1em plus 0.5em minus 0.4em\relax IEEE, 2023.

\bibitem{ju2024improving}
Y.~Ju, S.~Hu, S.~Jia, G.~H. Chen, and S.~Lyu, ``Improving fairness in deepfake detection,'' in \emph{WACV}, 2024.

\bibitem{lu2023attention}
Y.~Lu, B.~Kong, F.~Gao, K.~Cao, S.~Lyu, S.~Zhang, S.~Hu, Y.~Yin, and X.~Wang, ``Attention-driven tree-structured convolutional lstm for high dimensional data understanding,'' \emph{Frontiers in Physics}, vol.~11, p. 1095277, 2023.

\bibitem{xie2023attacking}
Y.~Xie, S.~Hu, X.~Wang, Q.~Liao, B.~Zhu, X.~Wu, and S.~Lyu, ``Attacking important pixels for anchor-free detectors,'' \emph{arXiv preprint arXiv:2301.11457}, 2023.

\bibitem{li2023ntire}
Y.~Li, Y.~Zhang, R.~Timofte, L.~Van~Gool, L.~Yu, Y.~Li, X.~Li, T.~Jiang, Q.~Wu, M.~Han \emph{et~al.}, ``Ntire 2023 challenge on efficient super-resolution: Methods and results,'' in \emph{CVPR}, 2023.

\bibitem{chen2023harnessing}
H.~Chen, P.~Zheng, X.~Wang, S.~Hu, B.~Zhu, J.~Hu, X.~Wu, and S.~Lyu, ``Harnessing the power of text-image contrastive models for automatic detection of online misinformation,'' in \emph{CVPR}, 2023.

\bibitem{hu2023rank}
S.~Hu, X.~Wang, and S.~Lyu, ``Rank-based decomposable losses in machine learning: A survey,'' \emph{IEEE Transactions on Pattern Analysis and Machine Intelligence}, 2023.

\bibitem{hu2022distributionally}
S.~Hu and G.~H. Chen, ``Distributionally robust survival analysis: A novel fairness loss without demographics,'' in \emph{Machine Learning for Health}.\hskip 1em plus 0.5em minus 0.4em\relax PMLR, 2022, pp. 62--87.

\bibitem{cao2017deformable}
X.~Cao, J.~Yang, J.~Zhang, D.~Nie, M.~Kim, Q.~Wang, and D.~Shen, ``Deformable image registration based on similarity-steered cnn regression,'' in \emph{MICCAI}.\hskip 1em plus 0.5em minus 0.4em\relax Springer, 2017.

\bibitem{yang2017fast}
X.~Yang, R.~Kwitt, M.~Styner, and M.~N. Quicksilver, ``Fast predictive image registration—a deep learning approach., 2017, 158,'' \emph{DOI: https://doi. org/10.1016/j. neuroimage}, vol.~8, pp. 378--396, 2017.

\bibitem{balakrishnan2019voxelmorph}
G.~Balakrishnan \emph{et~al.}, ``Voxelmorph: a learning framework for deformable medical image registration,'' \emph{TMI}, 2019.

\bibitem{sokooti2017nonrigid}
H.~Sokooti \emph{et~al.}, ``Nonrigid image registration using multi-scale 3d convolutional neural networks,'' in \emph{MICCAI}.\hskip 1em plus 0.5em minus 0.4em\relax Springer, 2017.

\bibitem{de2019deep}
B.~D. De~Vos \emph{et~al.}, ``A deep learning framework for unsupervised affine and deformable image registration,'' \emph{MIA}, 2019.

\bibitem{rohe2017svf}
M.-M. Roh{\'e} \emph{et~al.}, ``Svf-net: learning deformable image registration using shape matching,'' in \emph{MICCAI}.\hskip 1em plus 0.5em minus 0.4em\relax Springer, 2017.

\bibitem{tian2024unigradicon}
L.~Tian \emph{et~al.}, ``unigradicon: A foundation model for medical image registration,'' in \emph{MICCAI}.\hskip 1em plus 0.5em minus 0.4em\relax Springer, 2024.

\bibitem{wang2024advancing}
H.~Wang, Q.~Zhou, and Z.~Pan, ``Advancing medical image registration with the vision foundation model (vfm): A modular pre-trained framework,'' in \emph{2024 7th International Conference on Information Communication and Signal Processing (ICICSP)}.\hskip 1em plus 0.5em minus 0.4em\relax IEEE, 2024, pp. 1084--1088.

\bibitem{demir2024multigradicon}
B.~Demir, L.~Tian, H.~Greer, R.~Kwitt, F.-X. Vialard, R.~S.~J. Est{\'e}par, S.~Bouix, R.~Rushmore, E.~Ebrahim, and M.~Niethammer, ``Multigradicon: A foundation model for multimodal medical image registration,'' in \emph{International Workshop on Biomedical Image Registration}.\hskip 1em plus 0.5em minus 0.4em\relax Springer, 2024, pp. 3--18.

\bibitem{zhang2024challenges}
S.~Zhang and D.~Metaxas, ``On the challenges and perspectives of foundation models for medical image analysis,'' \emph{Medical image analysis}, vol.~91, p. 102996, 2024.

\bibitem{foret2020sharpness}
P.~Foret, A.~Kleiner, H.~Mobahi, and B.~Neyshabur, ``Sharpness-aware minimization for efficiently improving generalization,'' \emph{arXiv preprint arXiv:2010.01412}, 2020.

\bibitem{lin2024robust}
L.~Lin, S.~Papabathini, X.~Wang, and S.~Hu, ``Robust light-weight facial affective behavior recognition with clip,'' \emph{MIPR}, 2024.

\bibitem{lin2024robust1}
L.~Lin, I.~Amerini, X.~Wang, S.~Hu \emph{et~al.}, ``Robust clip-based detector for exposing diffusion model-generated images,'' \emph{MIPR}, 2024.

\bibitem{lin2024preserving}
L.~Lin, X.~He, Y.~Ju, X.~Wang, F.~Ding, and S.~Hu, ``Preserving fairness generalization in deepfake detection,'' in \emph{Proceedings of the IEEE/CVF Conference on Computer Vision and Pattern Recognition}, 2024, pp. 16\,815--16\,825.

\bibitem{davatzikos1997spatial}
C.~Davatzikos, ``Spatial transformation and registration of brain images using elastically deformable models,'' \emph{Computer Vision and Image Understanding}, vol.~66, no.~2, pp. 207--222, 1997.

\bibitem{rueckert1999nonrigid}
D.~Rueckert, L.~I. Sonoda, C.~Hayes, D.~L. Hill, M.~O. Leach, and D.~J. Hawkes, ``Nonrigid registration using free-form deformations: application to breast mr images,'' \emph{IEEE transactions on medical imaging}, vol.~18, no.~8, pp. 712--721, 1999.

\bibitem{beg2005computing}
M.~F. Beg, M.~I. Miller, A.~Trouv{\'e}, and L.~Younes, ``Computing large deformation metric mappings via geodesic flows of diffeomorphisms,'' \emph{International journal of computer vision}, vol.~61, pp. 139--157, 2005.

\bibitem{zhang2017frequency}
M.~Zhang, R.~Liao, A.~V. Dalca, E.~A. Turk, J.~Luo, P.~E. Grant, and P.~Golland, ``Frequency diffeomorphisms for efficient image registration,'' in \emph{Information Processing in Medical Imaging: 25th International Conference, IPMI 2017, Boone, NC, USA, June 25-30, 2017, Proceedings 25}.\hskip 1em plus 0.5em minus 0.4em\relax Springer, 2017, pp. 559--570.

\bibitem{vercauteren2009diffeomorphic}
T.~Vercauteren, X.~Pennec, A.~Perchant, and N.~Ayache, ``Diffeomorphic demons: Efficient non-parametric image registration,'' \emph{NeuroImage}, vol.~45, no.~1, pp. S61--S72, 2009.

\bibitem{avants2008symmetric}
B.~B. Avants, C.~L. Epstein, M.~Grossman, and J.~C. Gee, ``Symmetric diffeomorphic image registration with cross-correlation: evaluating automated labeling of elderly and neurodegenerative brain,'' \emph{Medical image analysis}, vol.~12, no.~1, pp. 26--41, 2008.

\bibitem{gu2018recent}
J.~Gu, Z.~Wang, J.~Kuen, L.~Ma, A.~Shahroudy, B.~Shuai, T.~Liu, X.~Wang, G.~Wang, J.~Cai \emph{et~al.}, ``Recent advances in convolutional neural networks,'' \emph{Pattern recognition}, vol.~77, pp. 354--377, 2018.

\bibitem{mok2020large}
T.~C. Mok and A.~C. Chung, ``Large deformation diffeomorphic image registration with laplacian pyramid networks,'' in \emph{Medical Image Computing and Computer Assisted Intervention--MICCAI 2020: 23rd International Conference, Lima, Peru, October 4--8, 2020, Proceedings, Part III 23}.\hskip 1em plus 0.5em minus 0.4em\relax Springer, 2020, pp. 211--221.

\bibitem{kim2021cyclemorph}
B.~Kim, D.~H. Kim, S.~H. Park, J.~Kim, J.-G. Lee, and J.~C. Ye, ``Cyclemorph: cycle consistent unsupervised deformable image registration,'' \emph{Medical image analysis}, vol.~71, p. 102036, 2021.

\bibitem{goodfellow2020generative}
I.~Goodfellow, J.~Pouget-Abadie, M.~Mirza, B.~Xu, D.~Warde-Farley, S.~Ozair, A.~Courville, and Y.~Bengio, ``Generative adversarial networks,'' \emph{Communications of the ACM}, vol.~63, no.~11, pp. 139--144, 2020.

\bibitem{kim2022diffusemorph}
B.~Kim, I.~Han, and J.~C. Ye, ``Diffusemorph: Unsupervised deformable image registration using diffusion model,'' in \emph{European conference on computer vision}.\hskip 1em plus 0.5em minus 0.4em\relax Springer, 2022, pp. 347--364.

\bibitem{ho2020denoising}
J.~Ho, A.~Jain, and P.~Abbeel, ``Denoising diffusion probabilistic models,'' \emph{Advances in neural information processing systems}, vol.~33, pp. 6840--6851, 2020.

\bibitem{vaswani2017attention}
A.~Vaswani, N.~Shazeer, N.~Parmar, J.~Uszkoreit, L.~Jones, A.~N. Gomez, {\L}.~Kaiser, and I.~Polosukhin, ``Attention is all you need,'' \emph{Advances in neural information processing systems}, vol.~30, 2017.

\bibitem{chen2022transmorph}
J.~Chen, E.~C. Frey, Y.~He, W.~P. Segars, Y.~Li, and Y.~Du, ``Transmorph: Transformer for unsupervised medical image registration,'' \emph{Medical image analysis}, vol.~82, p. 102615, 2022.

\bibitem{tian2023gradicon}
L.~Tian, H.~Greer, F.-X. Vialard, R.~Kwitt, R.~S.~J. Est{\'e}par, R.~J. Rushmore, N.~Makris, S.~Bouix, and M.~Niethammer, ``Gradicon: Approximate diffeomorphisms via gradient inverse consistency,'' in \emph{Proceedings of the IEEE/CVF Conference on Computer Vision and Pattern Recognition}, 2023, pp. 18\,084--18\,094.

\bibitem{greer2021icon}
H.~Greer, R.~Kwitt, F.-X. Vialard, and M.~Niethammer, ``Icon: Learning regular maps through inverse consistency,'' in \emph{Proceedings of the IEEE/CVF International Conference on Computer Vision}, 2021, pp. 3396--3405.

\bibitem{ronneberger2015u}
O.~Ronneberger, P.~Fischer, and T.~Brox, ``U-net: Convolutional networks for biomedical image segmentation,'' in \emph{Medical image computing and computer-assisted intervention--MICCAI 2015: 18th international conference, Munich, Germany, October 5-9, 2015, proceedings, part III 18}.\hskip 1em plus 0.5em minus 0.4em\relax Springer, 2015, pp. 234--241.

\bibitem{brock2021high}
A.~Brock, S.~De, S.~L. Smith, and K.~Simonyan, ``High-performance large-scale image recognition without normalization,'' in \emph{International conference on machine learning}.\hskip 1em plus 0.5em minus 0.4em\relax PMLR, 2021, pp. 1059--1071.

\bibitem{bahri2021sharpness}
D.~Bahri, H.~Mobahi, and Y.~Tay, ``Sharpness-aware minimization improves language model generalization,'' \emph{arXiv preprint arXiv:2110.08529}, 2021.

\bibitem{regan2011genetic}
E.~A. Regan, J.~E. Hokanson, J.~R. Murphy, B.~Make, D.~A. Lynch, T.~H. Beaty, D.~Curran-Everett, E.~K. Silverman, and J.~D. Crapo, ``Genetic epidemiology of copd (copdgene) study design,'' \emph{COPD: Journal of Chronic Obstructive Pulmonary Disease}, vol.~7, no.~1, pp. 32--43, 2011.

\bibitem{nevitt2006osteoarthritis}
M.~Nevitt, D.~Felson, and G.~Lester, ``The osteoarthritis initiative,'' \emph{Protocol for the cohort study}, vol.~1, p.~2, 2006.

\bibitem{van2012human}
D.~C. Van~Essen, K.~Ugurbil, E.~Auerbach, D.~Barch, T.~E. Behrens, R.~Bucholz, A.~Chang, L.~Chen, M.~Corbetta, S.~W. Curtiss \emph{et~al.}, ``The human connectome project: a data acquisition perspective,'' \emph{Neuroimage}, vol.~62, no.~4, pp. 2222--2231, 2012.

\bibitem{xu2016evaluation}
Z.~Xu, C.~P. Lee, M.~P. Heinrich, M.~Modat, D.~Rueckert, S.~Ourselin, R.~G. Abramson, and B.~A. Landman, ``Evaluation of six registration methods for the human abdomen on clinically acquired ct,'' \emph{IEEE Transactions on Biomedical Engineering}, vol.~63, no.~8, pp. 1563--1572, 2016.

\bibitem{bernard2018deep}
O.~Bernard \emph{et~al.}, ``Deep learning techniques for automatic mri cardiac multi-structures segmentation and diagnosis: is the problem solved?'' \emph{IEEE TMI}, 2018.

\bibitem{heimann2009comparison}
T.~Heimann \emph{et~al.}, ``Comparison and evaluation of methods for liver segmentation from ct datasets,'' \emph{IEEE TMI}, 2009.

\bibitem{hering2022learn2reg}
A.~Hering \emph{et~al.}, ``Learn2reg: comprehensive multi-task medical image registration challenge, dataset and evaluation in the era of deep learning,'' \emph{IEEE TMI}, 2022.

\bibitem{chen2023transmatch}
Z.~Chen, Y.~Zheng, and J.~C. Gee, ``Transmatch: a transformer-based multilevel dual-stream feature matching network for unsupervised deformable image registration,'' \emph{IEEE transactions on medical imaging}, vol.~43, no.~1, pp. 15--27, 2023.

\end{thebibliography}

\end{document}